%% file: main.tex
\documentclass[10pt,twocolumn,letterpaper]{article}

\usepackage{cvpr}
\usepackage{enumerate}
\usepackage{xcolor}
\usepackage{caption}
\usepackage{times}
\usepackage{epsfig}
\usepackage{graphicx}
\usepackage{amsmath}
\usepackage{amssymb}
\usepackage{wrapfig}
\usepackage[ruled, vlined]{algorithm2e}

\usepackage[pagebackref=true,breaklinks=true,letterpaper=true,colorlinks,bookmarks=false]{hyperref}

\cvprfinalcopy 
\def\cvprPaperID{1957} 

\newcommand*\samethanks[1][\value{footnote}]{\footnotemark[#1]}
\newcommand\blfootnote[1]{%
  \begingroup
  \renewcommand\thefootnote{}\footnote{#1}%
  \addtocounter{footnote}{-1}%
  \endgroup
}
\ifcvprfinal\fi
\begin{document}

\title{SBNet: Sparse Blocks Network for Fast Inference}

\author{
Mengye Ren\thanks{Equal contribution.}${\ \ }^{1,2}$, 
Andrei Pokrovsky\samethanks[1]${\ \ }^{1}$,
Bin Yang\samethanks[1]${\ \ }^{1,2}$, 
Raquel Urtasun${\ }^{1,2}$
\\
${}^{1}$Uber Advanced Technologies Group\\
${}^{2}$University of Toronto
\\
{\tt\small {\{mren3,andrei,byang10,urtasun\}@uber.com}}
}

\maketitle
\input{sections/abstract}
\input{sections/introduction}
\input{sections/related}
\input{sections/model}
\input{sections/experiments}
\input{sections/conclusion}


{\small
\bibliographystyle{ieee}
\bibliography{egbib}
}

\end{document}

%% file: sections/abstract.tex

\begin{abstract}

Conventional deep convolutional neural networks (CNNs) apply convolution operators uniformly in
space across all feature maps for hundreds of layers - this incurs a high computational cost for
real-time applications. For many problems such as object detection and semantic segmentation, we are
able to obtain a low-cost computation mask, either from {\it a priori} problem knowledge, or from a
low-resolution segmentation network. We show that such computation masks can be used to reduce
computation in the high-resolution main network. Variants of sparse activation CNNs have previously
been explored on small-scale tasks and showed no degradation in terms of object classification
accuracy, but often measured gains in terms of theoretical FLOPs without realizing a practical
speed-up when compared to highly optimized dense convolution implementations. In this work, we
leverage the sparsity structure of computation masks and propose a novel tiling-based sparse
convolution algorithm. We verified the effectiveness of our sparse CNN on LiDAR-based 3D object
detection, and we report significant wall-clock speed-ups compared to dense convolution without
noticeable loss of accuracy.
\blfootnote{Code available at \url{https://github.com/uber/sbnet}}

\end{abstract}

%% file: sections/introduction.tex
\section{Introduction}
Deep convolutional neural networks (CNNs) have led to major breakthroughs in many computer vision
tasks \cite{alexnet}. While model accuracy consistently improves with the number of
layers \cite{resnet}, as current standard networks use over a hundred convolution layers, the amount
of computation involved in deep CNNs can be prohibitively expensive for real-time applications such
as autonomous driving.

\begin{figure}[t!]
\centering
\includegraphics[width=0.95\columnwidth,trim={5cm 6.5cm 9cm 3cm}, clip]{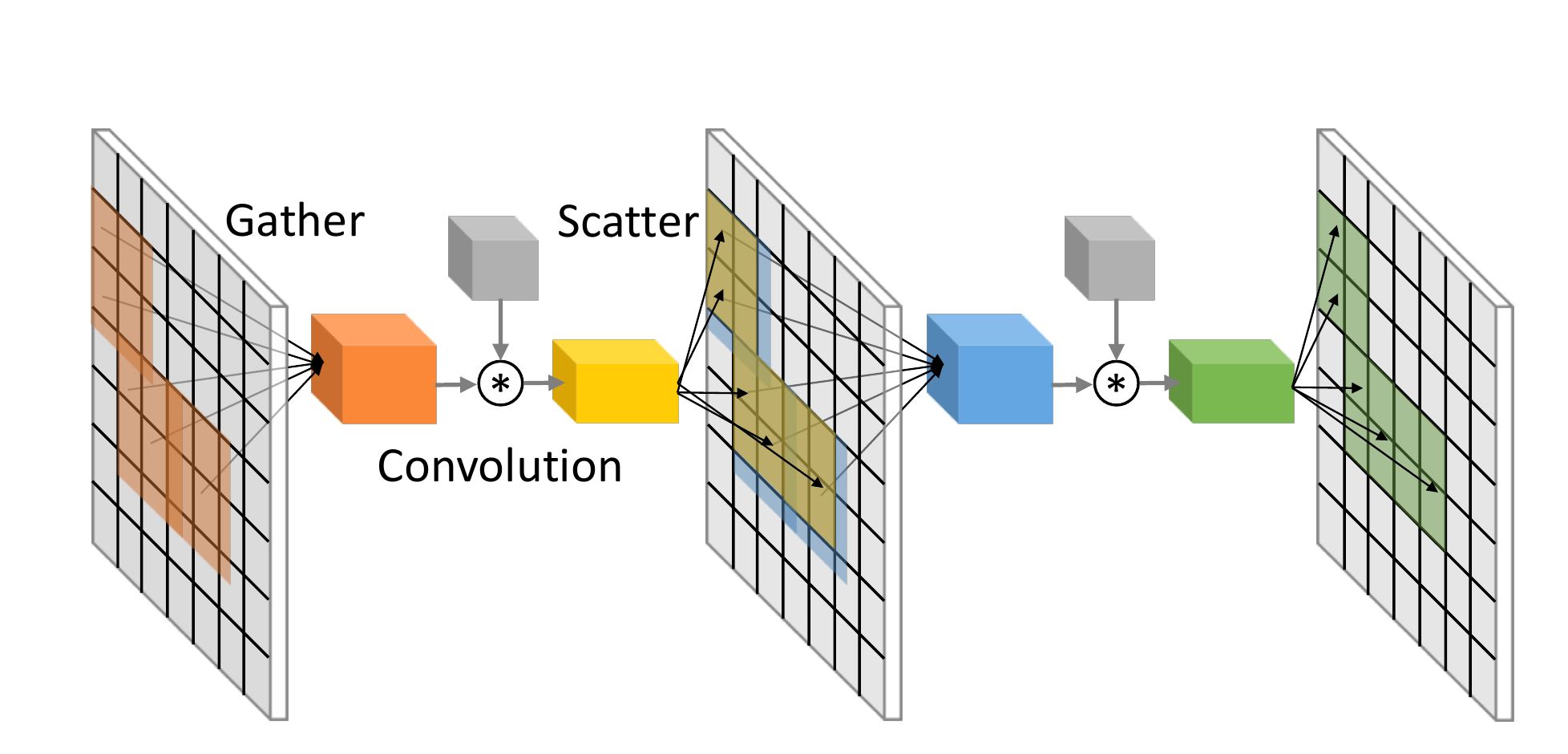}
\caption{Our proposed tiled sparse convolution module}
\label{fig:teaser}
\end{figure}

Spending equal amount of computation at all spatial locations is a tremendous waste, since spatial
sparsity is ubiquitous in many applications: in autonomous driving, only the areas on the road
matter for object detection; in video segmentation, only occluded and fast-moving pixels require
recomputation; in 3D object classification \cite{modelnet}, sparsity is directly encoded in the
inputs as voxel occupancy.  In these examples, spatial sparsity can be represented as binary
computation masks where ones indicate active locations that need more computation and zeros
inactive. In cases where such masks are not directly available from the inputs, we can predict them
in the form of visual saliency \cite{saliency} or objectness prior \cite{ron} by using another
relatively cheap network or even a part of the main network itself
\cite{figurnov2017adaptive,li2017nopixelequal}.

These binary computation masks can be efficiently incorporated into the computation of deep CNNs:
instead of convolving the input features at every location, we propose to use the masks to guide the
convolutional filters. Computation masks can also be considered as a form of attention mechanism
where the attention weights are binary. While most current uses of attention in computer vision have
been predominantly targeted at better model interpretability and higher prediction accuracy, our
work highlights the benefit of attentional inference speed-up.

In this work, we leverage structured sparsity patterns of computation masks and propose Sparse
Blocks Networks (SBNet), which computes convolution on a blockwise decomposition of the mask. We
implemented our proposed sparse convolution kernels (fragments of parallel code) on graphics
processing unit (GPU) and we report wall-clock time speed-up compared against state-of-the-art GPU
dense convolution implementations. Our algorithm works well with the popular residual network
(ResNet) architectures \cite{resnet} and produces further speed-up when integrated within a residual
unit.

Our sparse block unit can serve as a computational module in almost all deep CNNs for various
applications involving sparse regions of interest, bringing inference speed-up without sacrificing
input resolution or model capacity. We evaluate the effectiveness of our SBNet on LiDAR 3D object
detection tasks under a top-down bird's eye view, and we leverage both static road maps and dynamic
attention maps as our computation masks. We found SBNet achieves significant inference speedup
without noticeable loss of accuracy.

%% file: sections/related.tex
\section{Related work}

Sparse computation in deep learning has been extensively explored in the weights domain, where the
model size can be significantly reduced through pruning and low-rank decomposition
\cite{jaderberg2014lowrank,liu2015scnncvpr,han2015prune,
wen2016structsparse,li2017prunefilter,deeproots}. However it is not trivial to achieve huge speed-up
from sparse filters without loss of accuracy because a  single filter channel is rarely very close
to zero at every point. \cite{li2017prunefilter,channelprune} explored structured sparsity by
pruning an entire filter. Other forms of redundancies can also be leveraged such as weight
quantization \cite{incquant,limitquant}, teacher-student knowledge distillation \cite{distillation},
etc.

On the other end, in the activation domain, sparsity was also explored in various forms. Rectified
linear unit (ReLU) activations contain more than 50\% zero's on average and speed-up can be realized
on both hardware \cite{cnvlutin2} and algorithmic level \cite{reluspeedup}. Activation sparsity can
also be produced from a sparse multiplicative gating module \cite{moreisless}. In applications such
as 3D object classification, prior work also exploits structures in the sparse input patterns.
OctNet \cite{riegler2017octnet} introduces novel sparse high-resolution 3D representation for 3D
object recognition. Different from \cite{riegler2017octnet}, \cite{graham2017fbsparse} proposes a
generic valid sparse convolution operator where the input density mask is applied everywhere in the
network. As we will discuss later, while \cite{graham2017fbsparse} implements a generic convolution
operator, it is not suitable for moderately large input sizes.

When the inputs contain no structured sparsity, one can obtain dynamic computation masks during the
inference process over hundreds of layers. \cite{figurnov2017adaptive} learns to skip an adaptive
number of layers in ResNet for unimportant regions in object classification tasks. Similarly,
\cite{li2017nopixelequal} infers a pixel-wise mask for reweighting the computation in the context of
semantic segmentation. \cite{ron} predicts objectness prior heat maps during network inference for
more accurate object detection, but the heat maps do not help speed-up the inference process;
instead, the authors resort to downsampled inputs for faster inference. Given the vast availability
of those computation masks and heat maps during inference, our proposed sparse convolution operators
can be jointly applied to achieve major speedup gains on full resolution.

Sparse inference is beneficial to accuracy as the network focuses more of its computational
attention on useful activation patterns and ignores more of the background noise. For instance,
sparse batch normalization (BN) \cite{batchnorm,uhrig2017sparsebn} is invariant to input sparsity
level and outperforms regular BN in optical flow tasks. Here, we exploit the benefit of sparse BN
within our sparse residual units. Sparse convolution can also help increase the receptive field and
achieve better classification accuracy through perforated operations
\cite{figurnov2016perforatedcnn}.

Sparse computation masks are also related to the attention mechanism. Prior work applied visual
attention on convolutional features and obtained better model interpretability and accuracy on tasks
such as image captioning \cite{showattendtell}, visual question answering \cite{san,coattend}, etc.
However, unlike human attention which helps us reason visual scenes faster, these attentional
network structures do not speed up the inference process since the attention weights are dense
across the receptive field. Instead, we consider the simple case where the attention weights are
binary and explore the speed-up aspect of the attention mechanism in deep neural networks.

\paragraph{Comparison with \textit{im2col} based sparse convolution algorithms} Here we discuss the
main differences of our approach compared to popular sparse convolution algorithms based on matrix
lowering, as seen in \cite{liu2015scnncvpr,reluspeedup,moreisless}. These methods all use the same
type of matrix lowering which we refer as \textit{im2col}. Widely known in the implementation of
dense convolution in Caffe \cite{caffe}, \textit{im2col} gathers sliding windows of shape $k_H
\times k_W \times C$, where $k_H\times k_W$ is the filter window size and $C$ is the input channel
count. $B$ active windows are then reshaped into rows of a matrix of shape $B \times (k_H \times k_W
\times C)$ multiplied with a lowered filter matrix with shape $(k_H\times k_W \times C) \times
K$, where $K$ is the number of filters. This method is often faster than sparse matrix-vector
product due to contiguous memory access and better parallelism. However, these methods introduce
memory overhead and cannot leverage the benefits of Winograd convolution
\cite{winograd,lavin2016winograd}. Further, writing out the intermediate lowered results introduces
additional memory bandwidth overhead. \cite{graham2017fbsparse} designed a look-up table based data
structure for storing sparsity, but it is still slower compared to highly optimized Winograd
convolution. Our approach differs from \cite{graham2017fbsparse,li2017nopixelequal,reluspeedup} in
that we gather block-wise slices from tensors and maintain the tensor shape instead of lowering them
to vectors. Within each active block, we perform a \textit{regular} dense convolution and build on
top of a $2-3\times$ speedup from using Winograd convolution \cite{winograd,lavin2016winograd}
compared to general matrix-matrix multiplication (GEMM).

%% file: sections/model.tex
\section{SBNet: Sparse Blocks Network}

In this paper, we show that block sparsity can be exploited to significantly reduce the computational
complexity of convolutional layers in deep neural networks. Unlike previous work taking advantage of
unstructured sparsity, we show that our approach results in both theoretical and practical speed-up
without loss of accuracy. We observe that many input sources have structured sparsity that meshes
well with block sparsity - background pixels are likely to be surrounded by other background pixels.
It stands to reason that computations for entire spatial clumps or ``blocks'' of activations can be
skipped.

Block sparsity is defined in terms of a mask that can be known upfront from the input data domain
knowledge and {\it a priori} sparsity structure, or can be computed using lower cost operations. In
particular, we show the usefulness of our convolution algorithm on LiDAR object detection and we
exploit the sparsity from the road and sidewalk map mask as well as the model predicted foreground mask
at lower-resolution. For speed-up purposes, the same sparsity mask is reused for every layer in our
experiments, but it can also be computed from a different source per layer. In particular, at
different spatial scales within the network, we also use reduced spatial block sizes to better match
the granularity of spatial activations at that scale.

\begin{figure}
\centering
\includegraphics[width=0.95\columnwidth,trim={0.7cm 7cm 16cm 0cm}, clip]{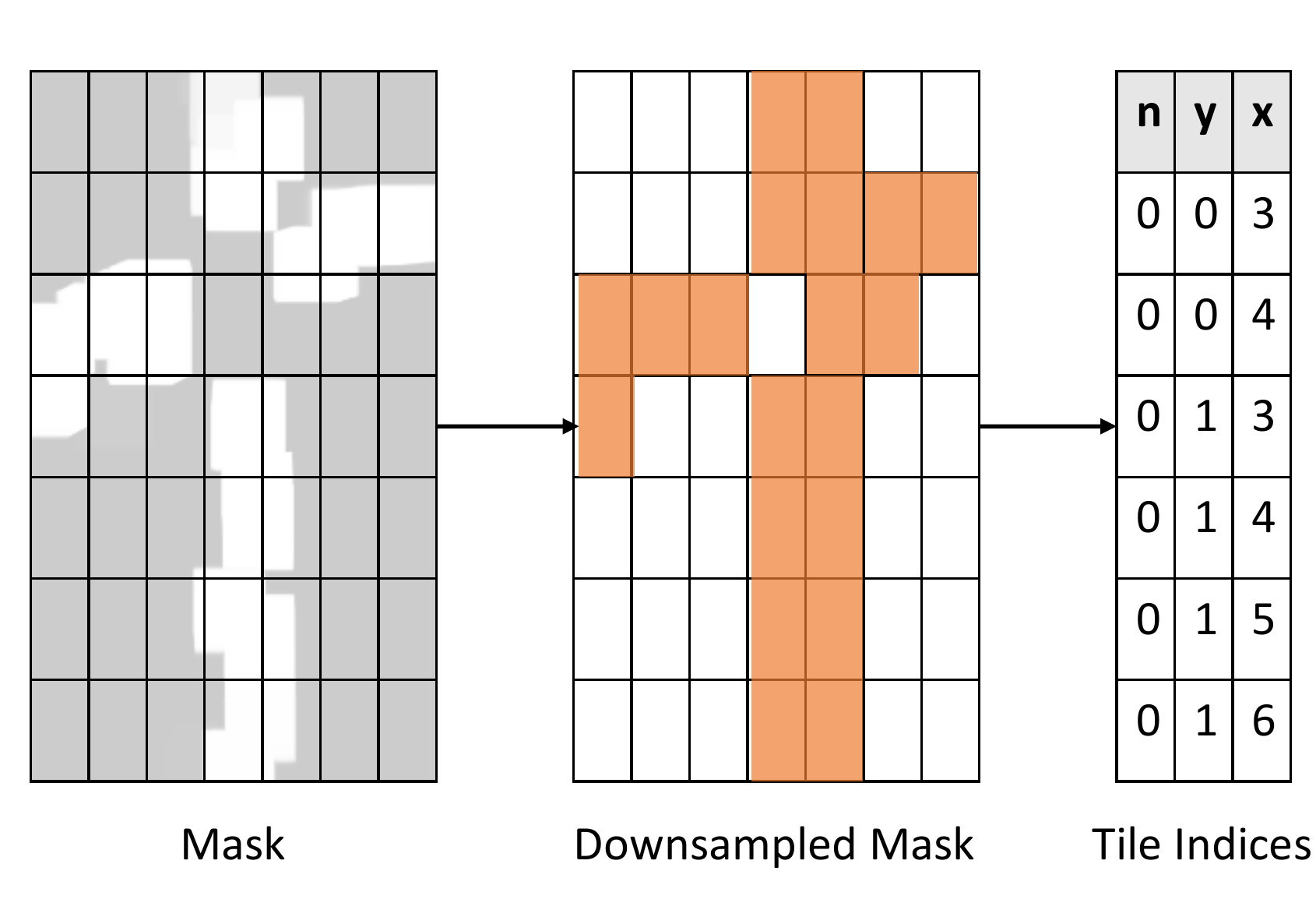}
\caption{Rectangular tiling for converting dense binary mask into sparse locations.}
\label{fig:reduce_mask}
\end{figure}

The input to our sparse convolution module is a dense binary mask. Just like other standard sparse
operations, we first need to extract a list of active location indices, which is named the
\textit{reduce mask} operation. Then, we would like to extract data from the sparse inputs at
specified locations and paste the computed results back to the original tensor. To summarize, there
are two major building blocks in our approach to sparse block-wise convolution:

\begin{enumerate}
    \item \textbf{Reduce mask to indices}: converts a binary mask to a list of indices, where each
index references the location of the corresponding $n$-dimensional block in the input tensor and in
our current implementation this is a 3-$d$ tuple (batch $n$, $y$-location, $x$-location) shared
across the channel dimension (see Figure~\ref{fig:reduce_mask}).
    \item \textbf{Sparse gather/scatter}: For gathering, we extract a block from the input
tensor, given the start location and the size of the $n$-d block. Scatter is the inverse operation
where we update the output tensor using previously gathered and transformed data.
\end{enumerate}

In this section, we first go over details of the above two building blocks, and then we introduce
a sparse blocks residual unit which groups several layers of computation into sparse blocks. Then
follows implementation details that are crucial to achieving a practical speed-up.

\subsection{Reduce mask to indices}

We start with a feature map of size $H \times W \times C$. We will demonstrate this for the case of
2D convolutions but our approach is applicable to higher dimensions. Let $M \in \{0,1\}^{H \times
W}$  be the binary mask representing the sparsity pattern. We would like to take advantage of
non-sparse convolution operations as they have been heavily optimized. With this in mind, we propose
to cover the non-zero locations with a set of rectangles.  Unfortunately, covering any binary shape
with a minimal number of rectangles is an NP-complete problem \cite{franklin1986rectanglecover}.
Furthermore, using rectangles of different shapes is hard to balance the computational load of
parallel processors. Therefore, we chose to have a uniform block size, so that the gathered blocks
can be batched together and passed into a single dense convolution operation.

\begin{figure}
\centering
\includegraphics[width=0.95\columnwidth,trim={0.8cm 7cm 19.5cm 0.3cm}, clip]{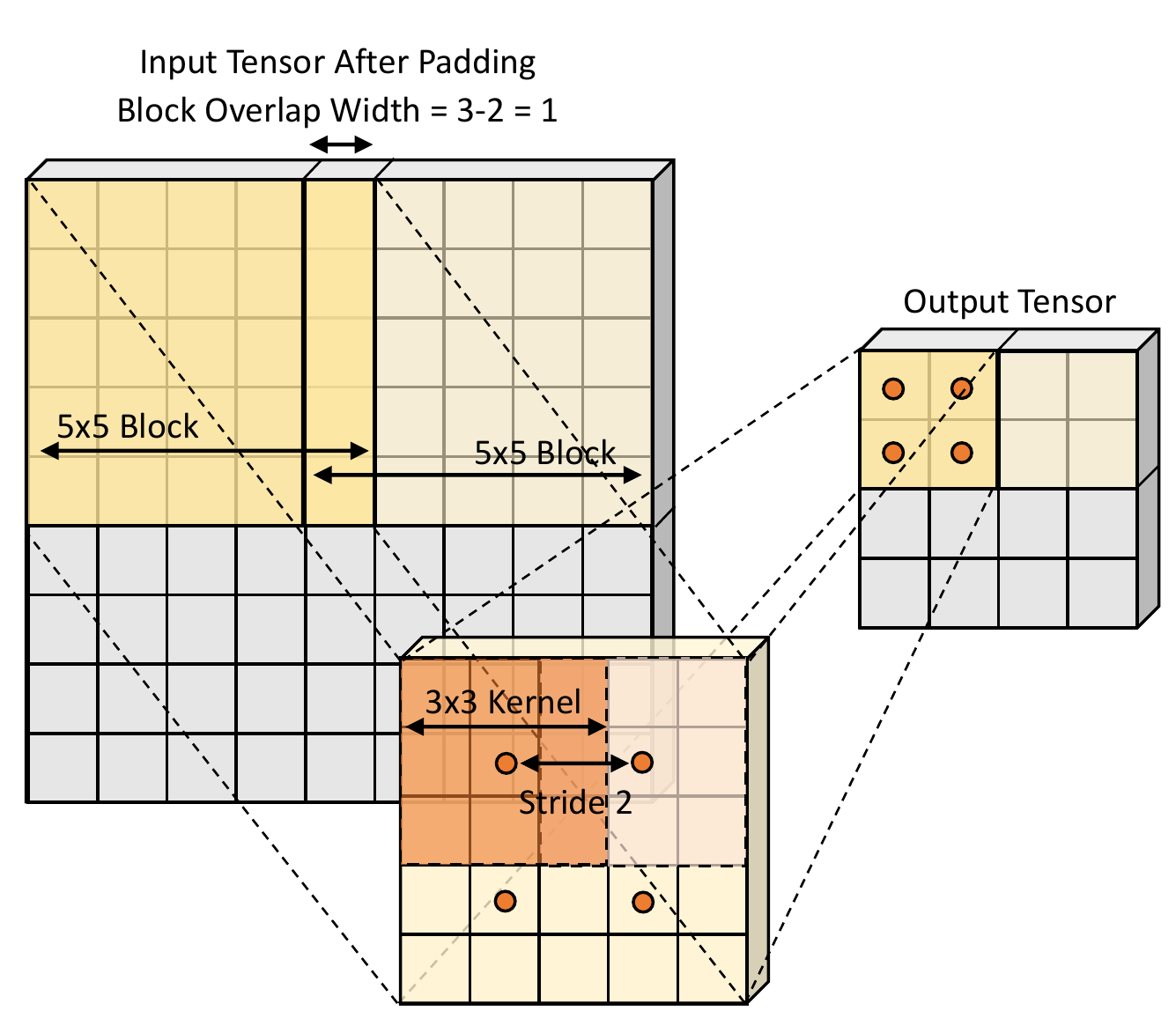}
\caption{A toy example with block size=$5\time5$, kernel size=$3\times3$, kernel strides=$2\times2$.
Block strides are computed as $k-s=3-2=1$.}
\label{fig:kernel_strides}
\end{figure}

In signal processing ``overlap-add'' and ``overlap-save'' are two standard partitioning schemes for
performing convolutions with very long input signals \cite{frerking1994dsp}. Our sparse tiling
algorithm is an instantiation of the ``overlap-save'' algorithm where we gather overlapping blocks,
but during the scatter stage, each thread writes to non-overlapping blocks so that the writes do not
require atomic locking. Knowing the block sizes and overlap sizes, we can perform a simple pooling
operation, such as maximum or average pooling followed by a threshold to downsample the input mask.
The resulting non-zero locations are the spatial block locations that we extract the patches from.
Figure~\ref{fig:kernel_strides} illustrates our tiling algorithm.

\subsection{Sparse gather/scatter}

Sparse gather/scatter operations convert the network between dense and sparse modes. Unlike regular
gather/scatter kernels that are implemented in deep learning libraries (e.g. \texttt{tf.gather\_nd,
tf.scatter\_nd}), our proposed kernels not only operate on dense indices but also expands spatially
to their neighborhood windows. Patch extracting operations (e.g. \texttt{tf.space\_to\_batch,
tf.batch\_to\_space}) also share some similarities with our approach but lack spatial overlap and
indexing capability. This input overlap is essential to producing the output that seamlessly
stitches the results of adjacent block convolutions in a way that is locally-equivalent to a dense
convolution on a larger block. Here, we introduce the technical details of our proposed gather and
scatter operations.

\paragraph{Gather kernel} Given a list of indices of size $[B,3]$, where $B$ is the number of
blocks, each has a tuple of ($n$, $y$, $x$) referencing the center location of the non-sparse
blocks, we then slice the blocks out of the 4-$d$ $N\times H\times W\times C$ input tensor using
$h\times w \times C$ slices, where $h$ and $w$ are the blocks' height and width, and stack the $B$
slices into a new tensor along the batch dimension, yielding a $B\times h \times w\times C$ tensor.

\paragraph{Scatter kernel} Scatter is an operation inverse to gather, reusing the same input mask
and block index list. The input to scatter kernel is a tensor of shape $B\times h'\times w'\times
C$. For a mini-network shown in Figure~\ref{fig:teaser}, $h'$ and $w'$ are computed according to the
output size reduction following a single unpadded convolution (also known as valid convolution).
This convolution is slotted between the scatter and gather operations. When this convolution has a
kernel size of $k_h\times k_w$ and strides $s_h\times s_w$, then, $h'=\frac{h-k_h+1}{s_h}$, and
$w'=\frac{w-k_w+1}{s_w}$. Figure~\ref{fig:kernel_strides} illustrates a toy example how the output
sizes are calculated.

\begin{figure}
\centering
\includegraphics[width=0.8\columnwidth,trim={0.0cm 0cm 17cm 0cm}, clip]{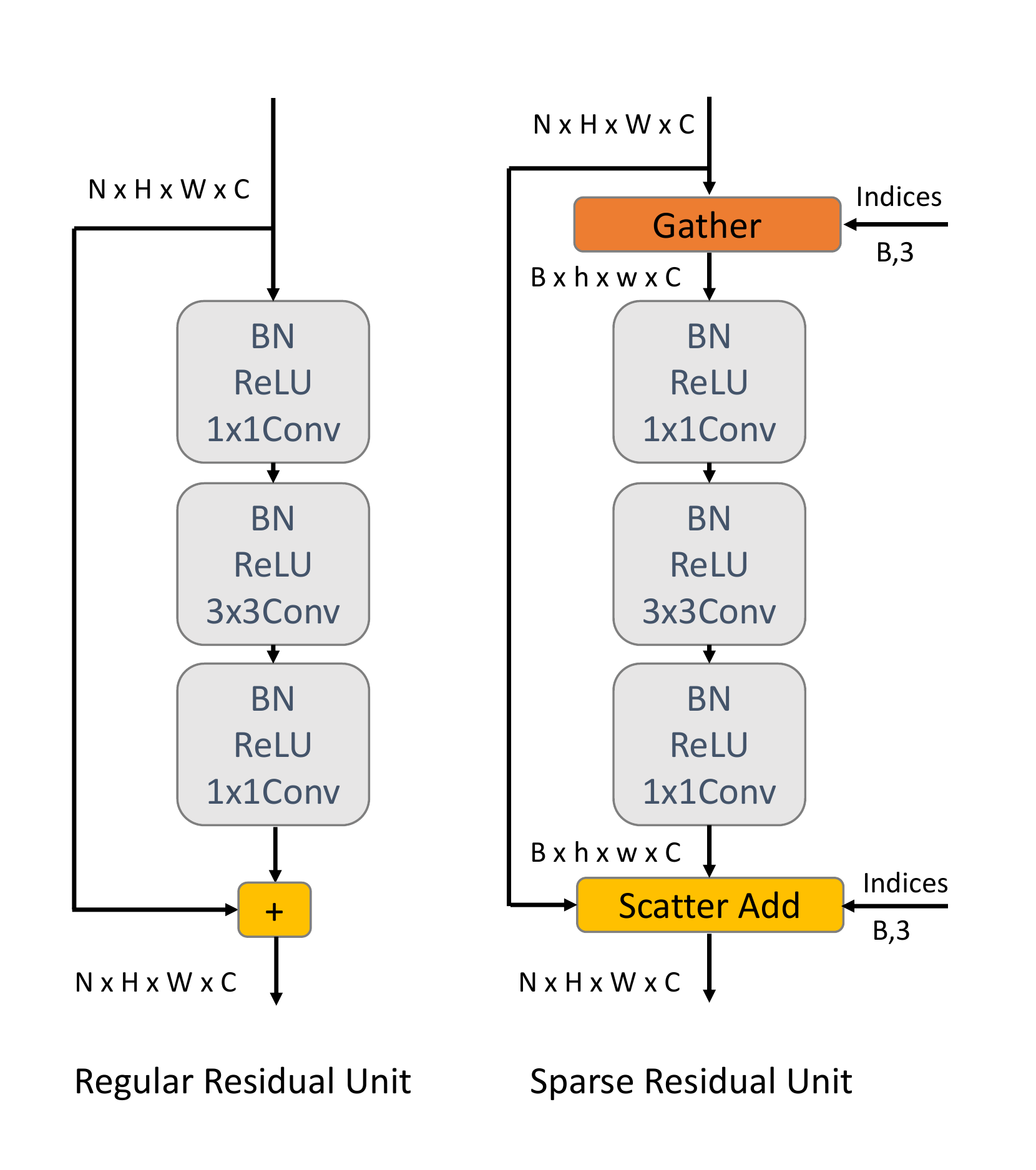}
\caption{A residual unit can be grouped into a sparse unit sharing one gather and scatter.}
\label{fig:sparse_res}
\end{figure}

\subsection{Sparse residual units} The ResNet architecture \cite{resnet} is widely used in many
state-of-the-art deep networks. Sparse residual units were previously explored using Valid Sparse
Convolution proposed in \cite{graham2017fbsparse}. Our proposed sparse blocks convolution also
integrates well with residual units. A single residual unit contains three convolutions, batch
normalization, and ReLU layers, all of which can be operated in sparse mode. The total increase in
receptive field of a residual unit is the same as a single $3\times3$ convolution. Therefore, all 9
layers can share a single pair of gathering and scattering operations without growing the overlap
area between blocks. In addition to the computation savings, \cite{uhrig2017sparsebn} showed that
batch-normalizing across non-sparse elements contributes to better model accuracy since it ignores
non-valid data that may introduce noise to the statistics. Figure~\ref{fig:sparse_res} shows a
computation graph of our sparse version of the residual unit.

\paragraph{End-to-end training of SBNet} is required since batch normalization (BN) statistics are
different between full-scale activations and dense-only activations. The gradient of a scatter
operation is simply the gather operation vice versa. When calculating the gradients of our
overlapping gather operation, the scatter needs to perform atomic addition of gradients on the edges
of overlapping tiles.

\subsection{Implementation details}

One of the major contributions of this work is an implementation of our block convolution algorithm
using custom CUDA kernels. As we will show in our experiments, this results in a significant
speed-up in terms of wall-clock time. This contrasts the literature, where only theoretical gains
are reported \cite{graham2017fbsparse}. In this section, we detail the techniques necessary to
achieve such speed-ups in practice.

\vspace{-0.2cm}
\paragraph{Fused downsample and indexing kernel} To minimize the intermediate outputs between
kernels, we fused the downsample and indexing kernels into one. Inside each tile, we compute a fused
max or average pooling operation followed by writing out the block index into a sequential index
array using GPU atomics to increment the block counter. Thus the input is a $N\times H\times W$
tensor and the output is a list of $B\times 3$ sparse indices referring to full channel slices
within each block.

\vspace{-0.2cm}
\paragraph{Fused transpose+gather and transpose+scatter kernels} When performing 2D spatial gather
and scatter, we favor $NHWC$ format because of channel memory locality: in $NHWC$ format, every
memory strip of size $w \times C$ is contiguous, whereas in $NCHW$ format, only strips of size $w$
are contiguous. Because cuDNN library runs faster with $NCHW$ data layout for convolutions and batch
normalization, our gather/scatter kernel also fuses the transpose from $NHWC$ to $NCHW$ tensor data
layout inside the same CUDA kernel. This saves a memory round-trip from doing additional transpose
operations and is instrumental in achieving a practical speed-up. 

\vspace{-0.2cm}
\paragraph{Fused scatter-add kernel for residual blocks} For ResNet architecture during inference,
the input tensor can be reused for output so that an extra memory allocation is avoided and there is
no need to wipe the output tensor to be all zeros. We implemented a fused kernel of 2D scatter and
addition, where we only update the non-sparse locations by adding the convolution results back to
the input tensor. 

%% file: sections/experiments.tex

\section{Experiments}

We validate our sparse blocks networks on our LiDAR 3D bird's eye view (BEV) detection benchmark
where the computation mask is available through offline road and sidewalk map information. In
addition to using a static map-based mask, we also explored using dynamic attention masks with
higher sparsity predicted by a small foreground segmentation network pretrained on dense box labels.
We investigate two key aspects of our proposed model: 1) inference speed-up compared to a dense deep
CNN detector; 2) change in detection accuracy brought by the use of sparse convolution.

\vspace{-0.2cm}
\paragraph{Experiment environments} For all of the experiments, we implemented and benchmarked in
TensorFlow 1.2.1 using cuDNN 6.0. Because TensorFlow by default uses $NHWC$ tensor format it incurs a
lot of overhead 
compared to cuDNN's preferred $NCHW$ format, 
we also implemented standard ResNet blocks in $NCHW$ for a fair comparison. 
To compare with the sub-manifold sparse convolution \cite{graham2017fbsparse}, we benchmark using
their released PyTorch implementation, using the same version of the cuDNN library. We use NVIDIA
GTX 1080Ti for the layerwise benchmark, and NVIDIA Titan XP for the full network benchmark.

\vspace{-0.2cm}
\paragraph{Choosing the optimal block sizes} Smaller block sizes produce higher mask matching
granularity at the expense of increased boundary overlap. Larger blocks have a lower percentage of
overlap, but depending on the feature map resolution, they are less usable due to their relative
size to the total size of the feature map. To achieve the maximum speed-up we perform a search sweep
over a range of block sizes to automatically pick the fastest-performing block decomposition.

\subsection{Datasets}
We used the following datasets for evaluating our LiDAR BEV detectors.

\begin{figure}
\centering
\fbox{\includegraphics[width=0.95\columnwidth,trim={0cm 0cm 0cm 0cm},
clip]{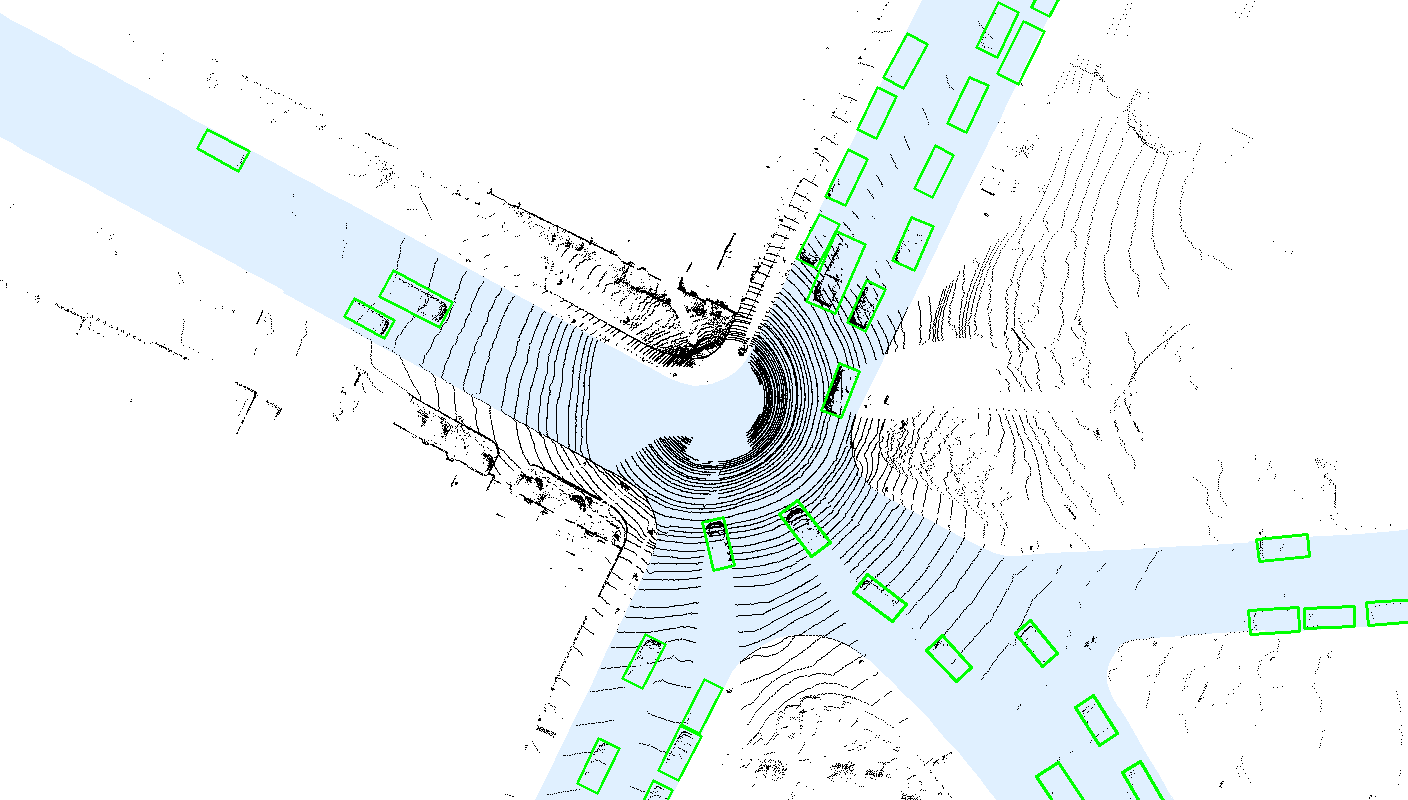}}
\caption{An example frame from our TOR4D LiDAR detection dataset. A single sweep over a region of
80m $\times$ 140.8m with a bird's eye view.  The road map is colored in blue, and ground-truth
detections are shown in green bounding boxes.}
\label{fig:tor4d}
\end{figure}

\paragraph{TOR4D}
Our internal TOR4D LiDAR detection dataset consists of 1,239,437 training frames, 5,979 validation
frames and 11,969 test frames. It also contains offline road map information, which can be directly
served as the computation mask without additional processing. Each frame contains LiDAR point cloud
sparse data for a region of 80m$\times$140.8m, with height ranging from -2m to 4m. We use
discretization bin size 0.1m$\times$0.1m$\times$0.2m. Two extra bins on the $z$-dimension are
designated to points outside the height range limits and one additional channel is used to encode
the LiDAR intensity. The input tensor of the detector is of size 800$\times$1408$\times$33. Each
frame has a corresponding crop of the road map, which is a top-down binary mask indicating which
pixels belong to the road (see Figure~\ref{fig:tor4d}).

\paragraph{KITTI}
To compare with other published methods, we also run experiments on the KITTI 2017 BEV benchmark
\cite{kitti}. The dataset consists of 7,481 training frames and 7,518 test frames. Each frame
contains a region of 80m$\times$70.4m, with height ranging from -3 to 1 m. We use discretization bin
size 0.1m$\times$0.1m$\times$0.2m. Two extra bins on the $z$-dimension are designated to points
outside the height range limits and one additional channel is used to encode the LiDAR intensity.
The input tensor of the detector is of size 800$\times$704$\times$23.

\subsection{Model}

\paragraph{3D object detector network} We adopt a fully convolutional detector architecture that
resembles \cite{focal}. Our model has a residual network backbone and one convolutional and two
upsampling layers with skip connections. For the residual backbone part, it has 2 initial
convolution layers (conv-1), followed by [3, 6, 6, 3] residual units per residual block (conv-2 -
conv-5), with channel depth [96, 192, 256, 384], and 16$\times$ downsampled activation size at the
top of the backbone network. Two extra upsampling (deconvolution) layers are appended to bring the
outputs back to 4$\times$ downsampled size, with skip connections from the outputs of conv-4 and
conv-3. Three branches of the outputs predict object classes, box sizes and orientations
respectively. Our sparse residual blocks and sparse convolutions are applied on all layers.

\vspace{-0.2cm}
\paragraph{Foreground mask network} To predict foreground computation masks, we adopt a Pyramid
Scene Parsing Network (PSPNet)  \cite{psp} on a ResNet-18 architecture \cite{resnet} at 8$\times$
downsampled input resolution. The network has no bottleneck layers and has one initial convolution
layer, followed by [2, 2, 2, 2] residual units per residual blocks, with channel depth [32, 64, 128,
256]. The network is trained to predict dilated dense box pixel labels.

\subsection{Experimental design}
We first run standalone layerwise speed-up tests, and we compare our approach with the theoretical
speed-up, i.e. 1/(1-sparsity), and the released implementation of sub-manifold sparse CNN
\cite{graham2017fbsparse} (``Sub-M''). Using the same activation size of our detector network, we
test the speed-up on three types of masks:
\begin{enumerate}[{1)}]
\itemsep0em 
\item \textit{Synthetic} masks generated using the top-left sub-region of input images to measure
the practical upper bound on speed-up.
\vspace{-0.07cm}
\item \textit{Road map} masks obtained from our offline map data in TOR4D.
\vspace{-0.07cm}
\item \textit{Predicted} masks obtained from the outputs of PSPNet.
\end{enumerate}
\vspace{-0.07cm}
We compare detection accuracy with two baselines: 
\begin{enumerate}[{1)}]
\itemsep0em 
\item \textit{Dense}: a dense network trained on all detection groundtruth.
\vspace{-0.07cm}
\item \textit{Dense w/ Road Mask}: a dense network trained on detection groundtruth within the road 
mask, i.e. treating regions outside the road as the ignore region. 
\end{enumerate}
\vspace{-0.07cm}
Our SBNets use computation masks from road and sidewalk maps and predicted masks, trained end-to-end
with the same number of training steps as the dense baselines. 
Detection accuracy is evaluated with on-road vehicles only.

\begin{figure*}[t]
\centering
\includegraphics[width=0.99\textwidth,trim={0cm 0cm 0cm 0cm}, clip]{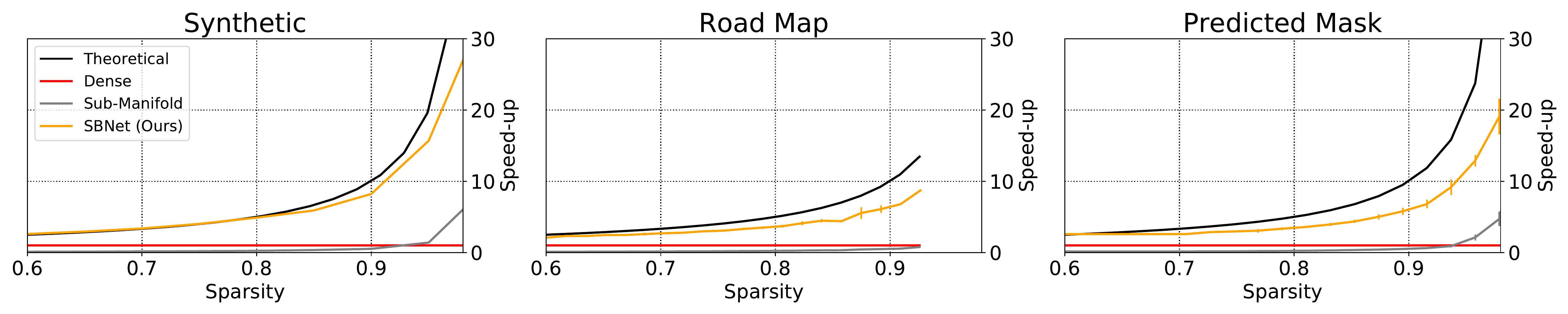}
\caption{
Residual block speed-up at resolution $400\times704$ (conv-2) for a range of sparsity level using
synthetic, road map, and predicted masks. Road masks do not have a full range of sparsity because
the dataset is collected on the road.}
\label{fig:res_speedup}
\end{figure*}

\begin{table}
\centering
\caption{Speed-up of a single $3\times3$ convolution on synthetic mask at 90\% sparsity. Theoretical
speed-up is 10.}
\label{tab:conv}
\begin{small}
\begin{tabular}{|c|c|c|c|}
\hline
Stage  & Size                     & Sub-M (\cite{graham2017fbsparse}) & SBNet (Ours) \\
\hline
conv-2 & 400$\times$704$\times$24 & 0.40$\times$                      &  3.39$\times$\\  
conv-3 & 200$\times$352$\times$48 & 0.75$\times$                      &  2.47$\times$\\
conv-4 & 100$\times$176$\times$64 & 0.28$\times$                      &  1.34$\times$\\
conv-5 & 50$\times$88$\times$96   & 0.13$\times$                      &  0.88$\times$\\
\hline
\end{tabular}
\end{small}
\end{table}

\begin{table}
\centering
\caption{Speed-up of residual units on synthetic masks at 90\% sparsity. Theoretical speed-up is 10.}
\label{tab:res_syn}
\resizebox{0.47\textwidth}{!}{
\begin{small}
\begin{tabular}{|c|c|c|c|c|}
\hline
Stage  & \#Units & Size                      & Sub-M (\cite{graham2017fbsparse}) & SBNet (Ours) \\
\hline
conv-2 & 3       & 400$\times$704$\times$96  & 0.52$\times$                      &  8.22$\times$\\  
conv-3 & 6       & 200$\times$352$\times$192 & 1.65$\times$                      &  6.27$\times$\\
conv-4 & 6       & 100$\times$176$\times$256 & 0.85$\times$                      &  3.73$\times$\\
conv-5 & 3       & 50$\times$88$\times$384   & 0.58$\times$                      &  1.64$\times$\\
\hline
\end{tabular}
\end{small}
}
\end{table}

\begin{figure}
\centering
\includegraphics[width=0.98\columnwidth,trim={0.0cm 0cm 0.0cm 0cm}, clip]{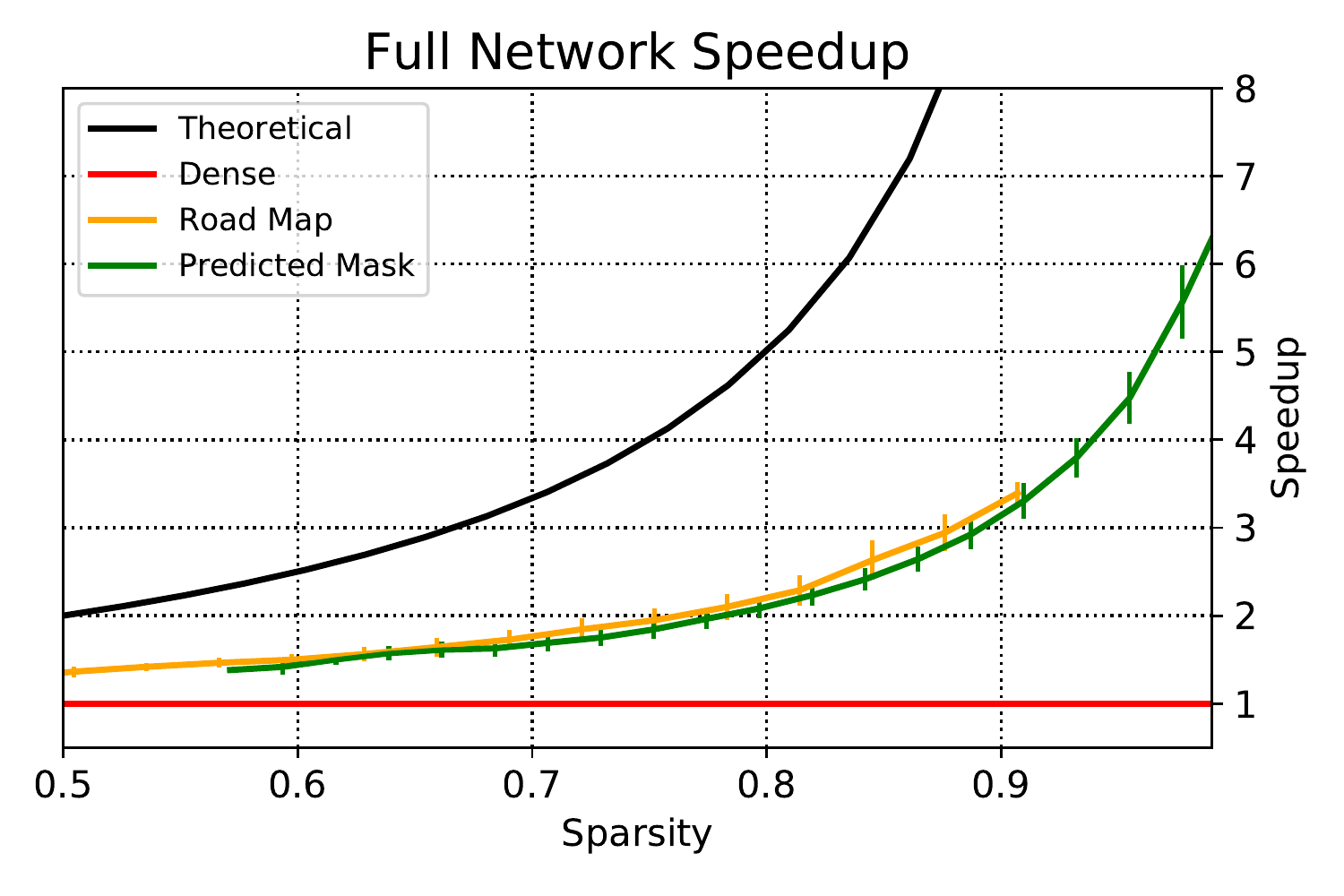}
\caption{
Full detector network speed-ups when using road map and predicted masks. An average speed-up in each
sparsity level is plotted with an error bar representing the standard deviation.}
\label{fig:full_speedup}
\end{figure}

\subsection{Results and Discussion}

Inference speed-ups for single convolution layers and residual blocks are listed in
Table~\ref{tab:conv}, \ref{tab:res_syn}, \ref{tab:res_road}, \ref{tab:res_psp}. For single
convolutions, our method achieves over 2 $\times$ speed-up for sparsity at 90\% at large
resolutions, whereas for residual units we obtain a significantly higher speed-up by grouping
multiple convolutions, BNs and ReLUs into a single sparse block sharing the sparse gather-transpose
and sparse scatter-transpose computation costs.

Notably, \cite{graham2017fbsparse} is slower than dense convolution on most activation sizes and
sparsity values, whereas our Sparse Blocks achieve much higher speed-up on large resolution sizes,
highlighting the practical contributions of our algorithm as increasing number of real-time
applications involve high-resolution inputs and outputs.

Figure~\ref{fig:res_speedup} plots speed-up vs. sparsity on conv-2 residual blocks, for three types
of different masks: \textit{synthetic}, \textit{road map}, and \textit{predicted}. Road maps and
predicted masks incur extra overhead compared to synthetic masks due to irregular shapes. Our method
significantly closes the gap between real implementations and the theoretical maximum and does not
slow down computation even at lower sparsity ratio such as 50 - 60\%, which is the typically the
least sparse road maps in our dataset. The computation masks output from the PSP network are 85 -
90\% sparse on average, bringing up the speed-up for all sparse layers (Table~\ref{tab:res_road}),
compared to using road masks (Table~\ref{tab:res_psp}), which are only 70 - 80\% sparse on average.

\begin{table}
\centering
\caption{Speed-up of residual units on road map masks at average 75\% sparsity. Theoretical speed-up
is 4.}
\label{tab:res_road}
\resizebox{0.47\textwidth}{!}{
\begin{small}
\begin{tabular}{|c|c|c|c|c|}
\hline
Stage  & \#Units & Size                      & Sub-M (\cite{graham2017fbsparse}) & SBNet (Ours) \\
\hline
conv-2 & 3       & 400$\times$704$\times$96  & 0.20$\times$                      & 3.05$\times$ \\  
conv-3 & 6       & 200$\times$352$\times$192 & 0.37$\times$                      & 2.15$\times$ \\
conv-4 & 6       & 100$\times$176$\times$256 & 0.50$\times$                      & 1.65$\times$ \\
conv-5 & 3       & 50$\times$88$\times$384   & 0.48$\times$                      & 1.14$\times$ \\
\hline
\end{tabular}
\end{small}
}
\end{table}

\begin{table}
\centering
\caption{Speed-up of residual units on PSPNet predicted masks at average 90\% sparsity. Theoretical 
speed-up is 10.}
\label{tab:res_psp}
\resizebox{0.47\textwidth}{!}{
\begin{small}
\begin{tabular}{|c|c|c|c|c|}
\hline
Stage  & \#Units & Size                      & Sub-M (\cite{graham2017fbsparse}) & SBNet (Ours) \\
\hline
conv-2 & 3       & 400$\times$704$\times$96  & 0.45$\times$                      & 5.21$\times$ \\  
conv-3 & 6       & 200$\times$352$\times$192 & 1.36$\times$                      & 3.25$\times$ \\
conv-4 & 6       & 100$\times$176$\times$256 & 0.77$\times$                      & 2.26$\times$ \\
conv-5 & 3       & 50$\times$88$\times$384   & 0.55$\times$                      & 1.32$\times$ \\
\hline
\end{tabular}
\end{small}
}
\end{table}

Table~\ref{tab:tor4d} reports detection accuracy on the TOR4D test set. We compare the road mask
version of SBNet with a dense baseline that has training loss masked with the road mask for a fair
comparison, since using road masks in the loss function hints learning more important regions. With
a significant 1.8$\times$ speedup, SBNet contributes to another 0.3\% gain in AP, suggesting that
sparse convolution and normalization layers during inference can be beneficial dealing with sparse
inputs. When using model predicted computation masks, we are able to reach 2.7$\times$ speedup, with
detection accuracy slightly below our dense baseline.

Comparison of our approach and other published methods on KITTI can be found in
Table~\ref{tab:kitti}. The dense detector baseline reached state-of-the-art performance in
``Moderate'' and ``Hard'' settings. The SBNet version of the detector achieves over 2.6$\times$
speed-up with almost no loss of accuracy. Including the cost of the mask network, our method is the
fastest among the top performing methods on the KITTI benchmark, an order of magnitude faster than
the published state-of-the-art \cite{mv3d}.

Detection results of our SBNet detector are visualized in Figure~\ref{fig:detect_results}. As shown,
PSPNet produces much sparser regions of interest compared to road maps while maintaining relatively
competitive detection accuracy. Many false negative instances have too few LiDAR points and are
difficult to be detected even by a dense detector.

Finally, we benchmark the computation overhead introduced by PSPNet in
Table~\ref{tab:mask_overhead}, which spends less than 4\% of the time of a full dense pass of the
detector network. SBNet and PSPNet combined together achieve 26.6\% relative gain in speed compared
to the Road Map counterpart. In addition to higher sparsity and speed-up, the predicted masks are
much more flexible in areas without offline maps.

\input{sections/figures_results}
\begin{table}
\centering
\caption{Speed-up \& detection accuracy of SBNet on the TOR4D dataset. AP at 70\% IoU.}
\label{tab:tor4d}
\resizebox{0.48\textwidth}{!}{
\begin{small}
\begin{tabular}{|c|c|c|c|c|}
\hline
Model       & Train Loss &  Sparsity      & Avg. Speed-up     & AP         \\
\hline 
Dense       & Road Mask  &  0\%           & 1.0$\times$       & 75.70       \\
SBNet +Road & Road Mask  & 70\%           & 1.78$\times$      & \bf{76.01}  \\
\hline
Dense       & No Mask    &  0\%           & 1.0$\times$       & \bf{73.28}  \\
SBNet +PSP  & PSP Mask   & 86\%           & \bf{2.66$\times$} & 73.01       \\
\hline
\end{tabular}
\end{small}
}
\end{table}

\begin{table}[h]
\centering
\caption{KITTI Bird's Eye View (BEV) 2017 Benchmark}
\begin{small}
\begin{tabular}{|c|c|c|c|c|}
\hline
Model               & Moderate    & Easy       & Hard       & Avg. Time     \\
\hline
DoBEM \cite{bevcnn} & 36.95       & 36.49      & 38.10      & 600 ms        \\
3D FCN \cite{3dfcn} & 62.54       & 69.94      & 55.94      & $>$ 5 s       \\
MV3D \cite{mv3d}    & 77.00       & \bf{85.82} & 68.94      & 240 ms        \\
\hline
Dense               & \bf{77.05}  & 81.70      &\bf{72.95}  & 47.3 ms       \\
SBNet               & 76.79       & 81.90      & 71.40      & \bf{17.9 ms}  \\
\hline
\end{tabular}
\vspace{-0.2cm}
\label{tab:kitti}
\end{small}
\end{table}

\begin{table}
\centering
\caption{Mask network computation overhead}
\label{tab:mask_overhead}
\begin{small}
\begin{tabular}{|c|c|c|}
\hline
Network        & Resolution      & Time (ms)  \\
\hline
Dense          & $800\times1408$ & 88.0       \\
\hline
SBNet +Road    & $800\times1408$ & 49.5       \\
SBNet +PSP     & $800\times1408$ & 33.1       \\
\hline
PSPNet         & $100\times176$  & 3.2        \\
\hline
\end{tabular}
\end{small}
\end{table}

%% file: sections/figures_results.tex
\newcommand{\resultsImg}[1] {
\includegraphics[trim={0 0cm 0 0cm},clip,width=0.33\textwidth]{./figures/output_vis-dense-mask/#1_vis.png} &
\includegraphics[trim={0 0cm 0 0cm},clip,width=0.33\textwidth]{./figures/output_vis-sparse/#1_vis.png} &
\includegraphics[trim={0 0cm 0 0cm},clip,width=0.33\textwidth]{./figures/output_vis-sparse-net3/#1_vis.png}
}

\begin{figure*}[t]
\centering
\resizebox{\textwidth}{!}{
\def\arraystretch{1.8}
\begin{tabular}{ccc}
Dense & SBNet +Road & SBNet +PSP \\
\hline
\resultsImg{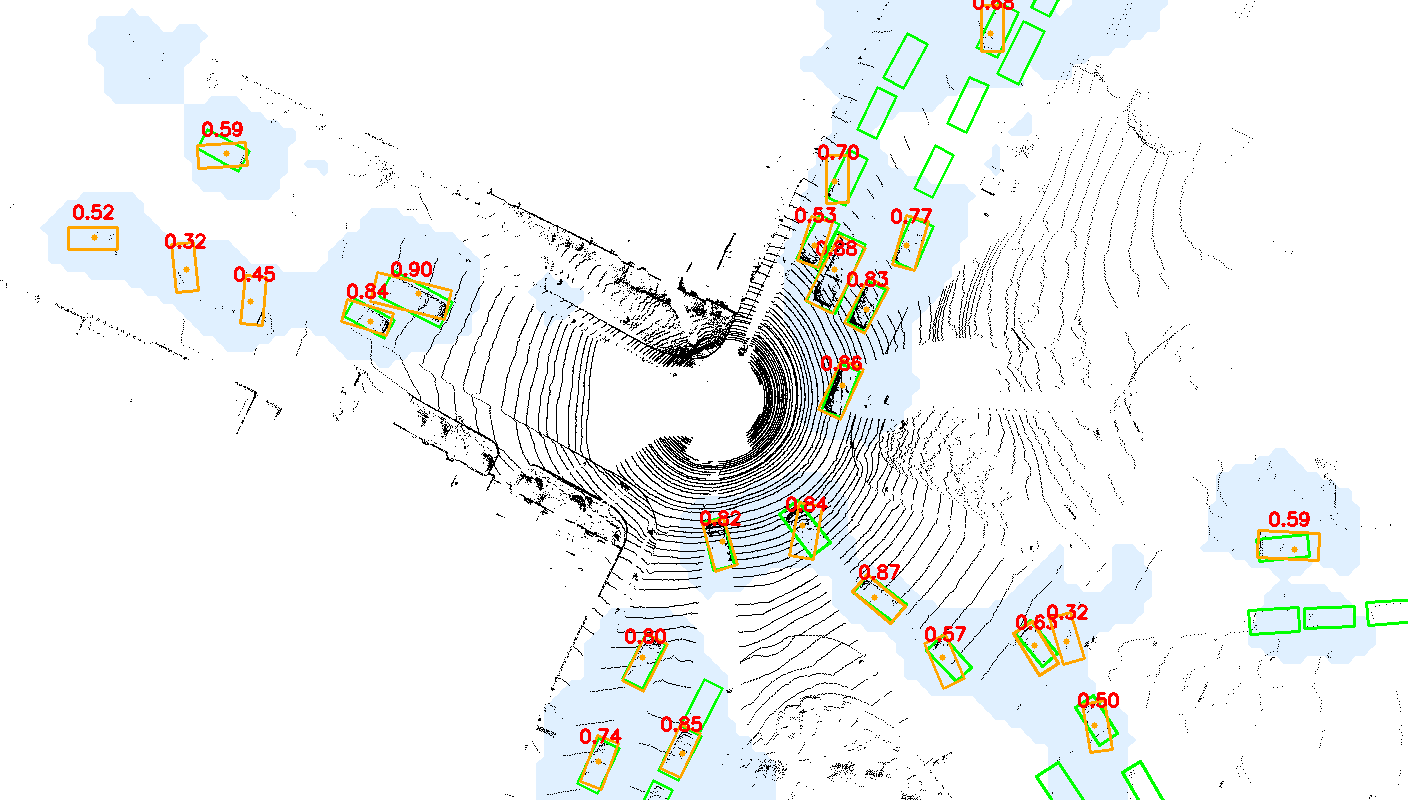}\\
\hline
\resultsImg{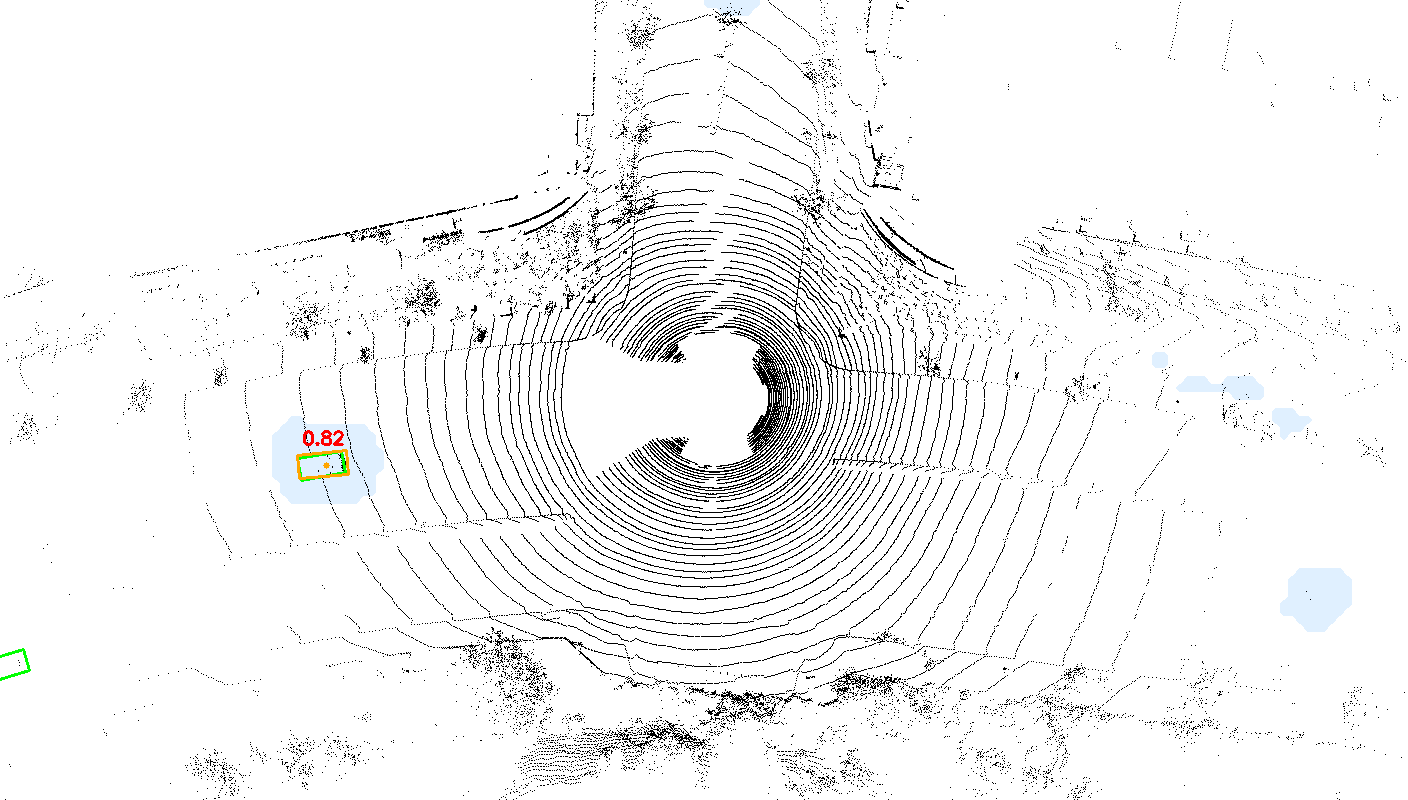}\\
\hline
\resultsImg{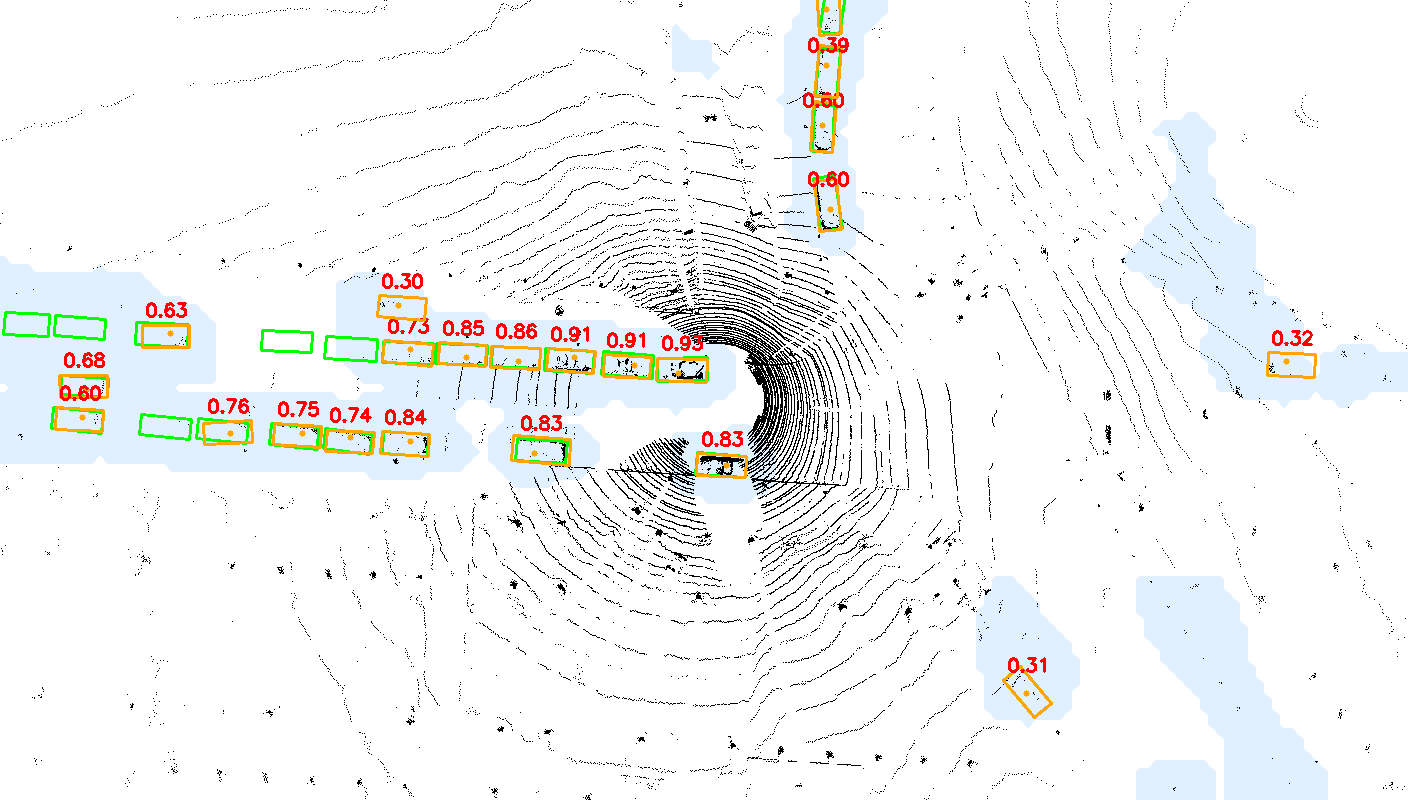}\\
\end{tabular}
}
\caption{A bird's eye view of our 3D vehicle detection results. Green boxes denote groundtruth and
orange denote outputs. Blue regions denote sparse computation masks. Visit
\url{https://eng.uber.com/sbnet} for a full video.}
\label{fig:detect_results}
\end{figure*}

%% file: sections/conclusion.tex
\section{Conclusion and Future Work}
In this work, we introduce the Sparse Blocks network which features fast convolution computation
given a computation mask with structured sparsity. We verified significant wall-clock speed-ups
compared to state-of-the-art dense convolution implementations. In LiDAR 3D detection
experiments, we show both speed-up and improvement in detection accuracy using road map masks, and
even higher speed-up using model predicted masks while trading off a small amount of accuracy. We
expect our proposed algorithm to achieve further speed-up when used jointly with other orthogonal
methods such as weights pruning, model quantization, etc. As future work, sparse blocks can be
extended to a combination of different rectangle shapes (c.f. OctNet \cite{riegler2017octnet}) to
get fine-grained mask representation, which can speed up inference with multi-scaled reasoning.